A Robot for Nondestructive Assay of Holdup Deposits in Gaseous Diffusion Piping – 19504

Heather Jones *, Siri Maley *, Mohammadreza Mousaei *, David Kohanbash *, Warren Whittaker *,
James Teza *, Andrew Zhang *, Nikhil Jog*, William Whittaker *
* Carnegie Mellon University

**ABSTRACT**

Miles of contaminated pipe must be measured, foot by foot, as part of the decommissioning effort at deactivated gaseous diffusion enrichment facilities. The current method requires cutting away asbestos-lined thermal enclosures and performing repeated, elevated operations to manually measure pipe from the outside. The RadPiper robot, part of the Pipe Crawling Activity Measurement System (PCAMS) developed by Carnegie Mellon University and commissioned for use at the DOE Portsmouth Gaseous Diffusion Enrichment Facility, automatically measures U-235 in pipes from the inside. This improves certainty, increases safety, and greatly reduces measurement time.

The heart of the RadPiper robot is a sodium iodide scintillation detector in an innovative disc-collimated assembly. By measuring from inside pipes, the robot significantly increases its count rate relative to external through-pipe measurements. The robot also provides imagery, models interior pipe geometry, and precisely measures distance in order to localize radiation measurements. Data collected by this system provides insight into pipe interiors that is simply not possible from exterior measurements, all while keeping operators safer.

This paper describes the technical details of the PCAMS RadPiper robot. Key features for this robot include precision distance measurement, in-pipe obstacle detection, ability to transform for two pipe sizes, and robustness in autonomous operation. Test results demonstrating the robot's functionality are presented, including deployment tolerance tests, safeguarding tests, and localization tests. Integrated robot tests are also shown.

**INTRODUCTION**

Safe, efficient cleanup of legacy uranium enrichment processing sites poses a significant challenge to the United States Department of Energy (DOE), and achieving "cold and dark" status at gaseous diffusion facilities is critical to the nuclear decontamination and deactivation (D&D) mission. Each gaseous diffusion plant contains hundreds of miles of pipe that must be inspected for U-235 holdup deposits and certified safe before the plant can be demolished. Pipes containing less U-235 than the criticality incredible (CI) threshold are considered safe to demolish in place, but pipes containing unsafe amounts of fissile material require costly removal and cleaning before disposal. The current nondestructive assay (NDA) method requires measurement of radiation from the pipe exterior to estimate the U-235 content per foot of pipe. This process is slow, labor-intensive, error-prone, and poses significant occupational hazards to plant personnel. A new turnkey system of robotic inspection, with automatic analysis, reporting, and archiving, is developed herein to perform a high-cadence NDA from within process piping with detector and sensors that directly view the holdup deposits.

In Situ Object Counting System (ISOCS) and Holdup Measurement System 4 (HMS4) are examples of established radiometric assay techniques for enrichment piping [1,2]. They require manual deployment at elevation, suffer gamma attenuation through pipe walls, require long counting time, need approximate modeling, and have shortfalls associated with transcription and human interpretation. More recently, a robotic NDA approach from inside pipes using an innovative disc collimation method was demonstrated, enabling faster measurement speeds, automated correlation of radiation measurements to along-pipe positions, and collection of rich auxiliary data [3,4]. This work builds from that proof-of-concept to production prototype for the Pipe Crawling Activity Measurement System (PCAMS) RadPiper robot.

As in [3], a now-improved autonomous robotic crawler carries a disc-collimated detector assembly through pipes and acquires radiometric, visual, and geometric data. In this work, the robot also includes safeguarding that builds three-dimensional models of the pipe ahead to avoid a wide range of obstacles. Robust autonomy responds to that 3D safeguarding mapper, point-lasers used as secondary safeguards, and various on-board health and status checks to ensure that the robot can safely return to its launch point. The new robot includes quality control checks using an onboard radioactive source. The robot and supporting hardware have the ability to configure for inspection of two different pipe diameters, 30- and 42-inch, for traversal at 10 and 6 feet per minute, respectively. (This is the equivalent 3 meters and 2 meters per minute in 76 and 107 cm pipe, respectively). An additional rangefinding sensor and improved software enable better localization of all sensor measurements.

The combination of automated RadPiper robots (Fig. 1) to inspect pipe interiors and the accompanying analysis software to automatically analyze data and generate official NDA reports enables speed, safety, accuracy, and economy unachievable by prior or competing approaches. The automated system provides full end-to-end NDA data logging, analysis, visualization, and long-term storage while minimizing risk to personnel and concentrating human efforts on key automatically-detected edge cases.

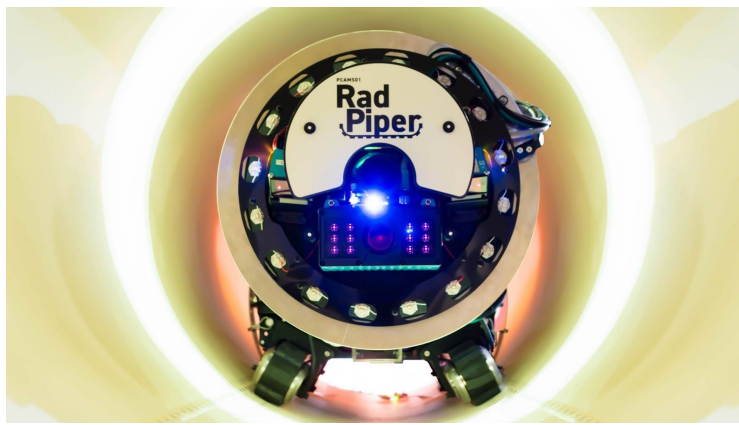

Fig. 1. PCAMS RadPiper in a 30-inch pipe.

While the manual inspection method requires approximately five man-hours per foot of pipe, RadPiper is capable of inspecting a 150-foot section of pipe in approximately one hour and generating a full, approvable NDA report in less than that. This entire pipeline was hot tested by DOE Portsmouth technicians and analysts in July 2018 (Fig. 2). Thus far two production prototype RadPiper units have been designed, manufactured, documented, and tested by Carnegie Mellon University and Portsmouth.

This paper concentrates on details of the autonomous RadPiper robot (Fig. 3), beginning with a brief discussion of radiometry and quality control (detailed in [5]), followed by a description of non-radiometric sensors for NDA, localization, and robot autonomy. The robot's structures, mechanisms, electronics, and computing are then described with respect to performance objectives such as radiometric certainty, data quality, and safety. A section on autonomy discusses the robot's executive software and its safeguarding. A section on localization describes determination of precise robot positions (synchronization of positions with radiation measurements is detailed in [6]). The discussion concludes with results of feature-level and integrated performance testing, with hot testing results available in [7] and calibration testing in [5]. Fig. 3 shows an exploded view of RadPiper, highlighting relevant sensors and structures.

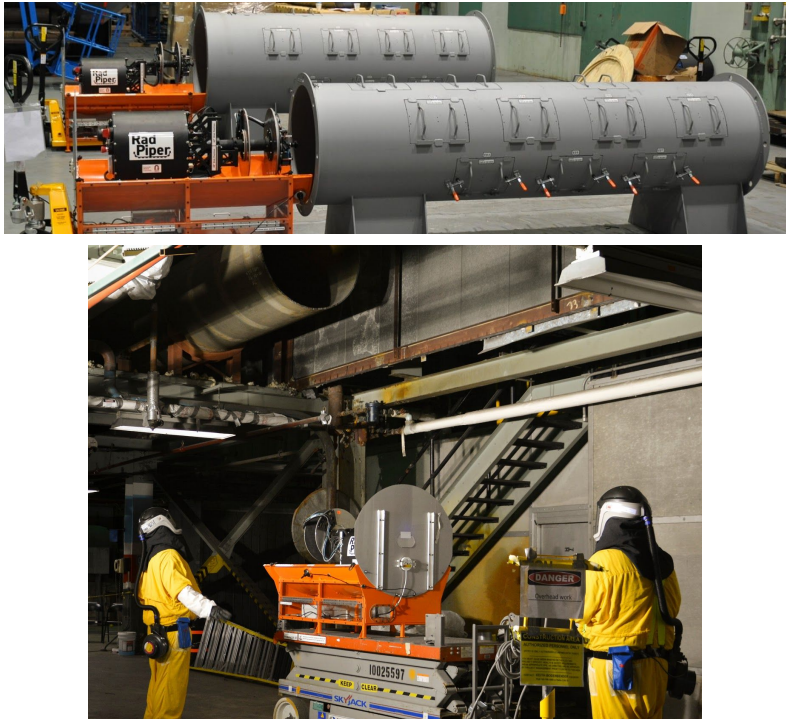

Fig. 2. (Top) Phase 1 RadPiper PCAMS robots undergoing testing at DOE Portsmouth. (Bottom) A RadPiper robot deploying in a Portsmouth process pipe.

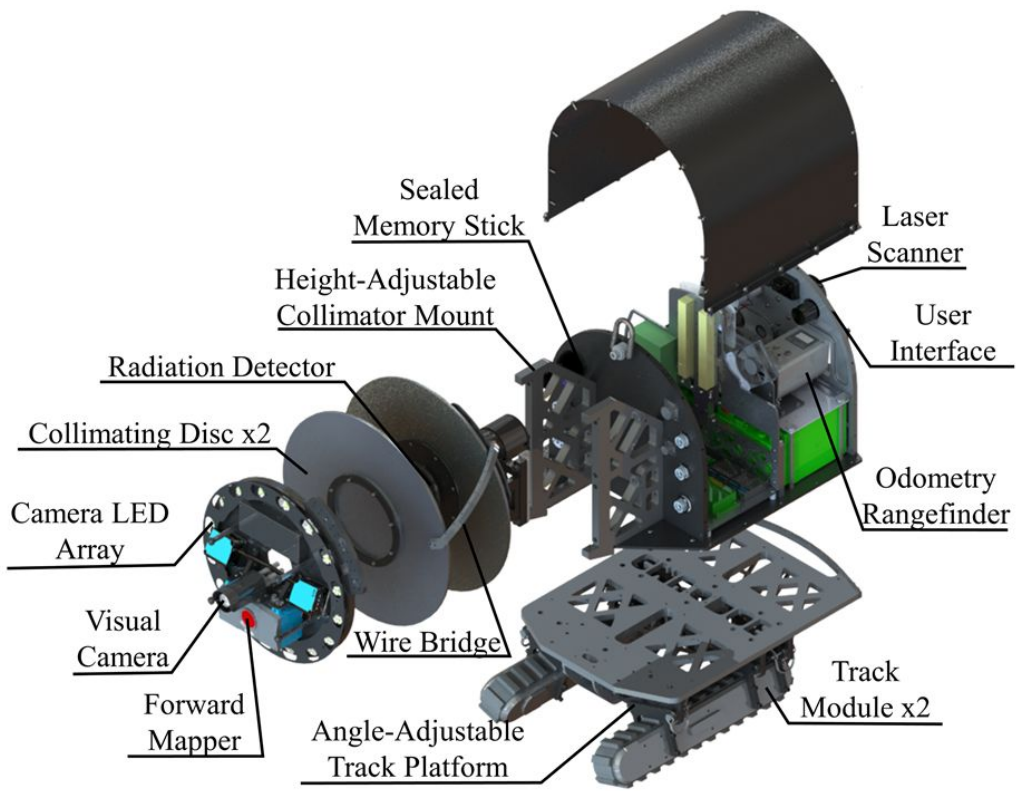

Fig. 3. Exploded view of PCAMS RadPiper robot.

## RADIOMETRY AND QUALITY CONTROL

The speed and counting certainty with which PCAMS conducts NDA is made possible by the invention and integration of a disc-collimated gamma detector [8]. This novel sensor places a 2x2 NAIS Sodium Iodide Scintillation Detector (5 cm long, 5 cm diameter) detector between two aluminum-backed lead discs, restricting the crystal's field of view to a axisymmetric cylindrical annulus. The RadPiper robot centers this collimated gamma detector along the axis of a 30- or 42-inch pipe and continuously traverses in this orientation, allowing the detector to rapidly view a continuously-moving annulus of contaminated pipe wall with full radial symmetry.

This collimator geometry is illustrated in schematic and three-dimensional view in Fig. 4. Note that for practical purposes, the main annulus of view (blue) is appended by two known-length tail regions (red) due to the finite sizes of the detector crystal and collimating discs. The PCAMS collimated detector and radiometric method are discussed in depth in [5].

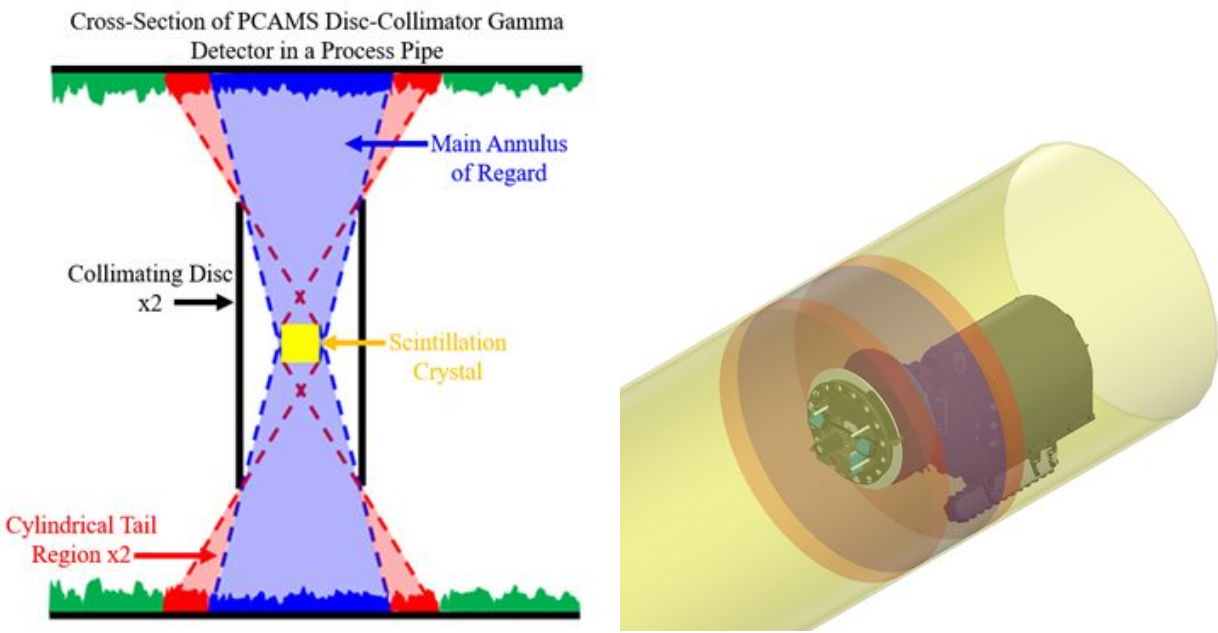

Fig. 4. (Left) schematic view and (right) isometric rendering of RadPiper's disc collimator in a gas diffusion pipe, showing the gamma detector's primary annulus of regard in blue and its tail regions in red.

Unlike the prior prototype robot with disc collimation [3], these production prototypes contain an onboard Am-241 check source. Quality control data are collected both before and after each pipe run. Because the check source is known, the robot automatically compares the detected and expected Am-241 peak location, width, and counts in its region of interest. If these bounds are exceeded, RadPiper automatically stops its run and alerts the NDA technician (via wireless tablet) to investigate the problem.

In addition, repeated traversal over holdup deposit opens the potential for uranyl fluoride contaminating the robot in a way that would interfere with radiometric measurements. RadPiper thus automatically compares the counts in its U-235 (186 keV) region of interest collected during its pre-run and post-run quality control checks. If the detected activity is different (beyond 3-sigma uncertainty on the measurement), then the NDA technician is notified of a potential contamination issue and follows

PCAMS procedures for cleaning and retesting the system.

## AUXILIARY SENSING

In addition to a radiation detector, RadPiper carries a number of sensors that are used to ensure that model assumptions for radiation measurements are not violated, (or to enable data to be flagged when they are), and to help analysts interpret radiation measurements. A fisheye camera captures images looking forward down a pipe for situational awareness of technicians and analysts. Illumination for imaging is provided by sixteen onboard white LEDs (see Fig. 5).

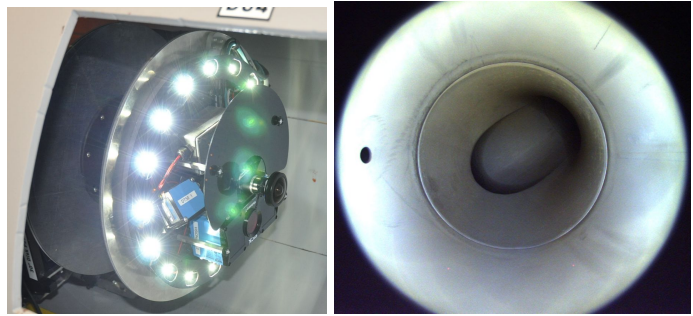

Fig. 5. (Left) RadPiper's front module, showing illuminated LEDs and fisheye camera. (Right) Camera image taken inside a process pipe as it approaches the arm of a swept T joint. An instrument port is visible on the left.

A spinning 2-dimensional triangulation laser scanner (RPLIDAR A3) is mounted on the back of the robot to gather geometric profiling data. It is also used for flagging pipe segments for which deposit geometry would cause excessive attenuation, (resulting in a radiometric underestimate), or when there is hole or fitting that breaks the round-pipe assumption used in radiometry. This sensor is also used to register robot position to the start of a pipe, so that reported measurements are relative to a real-world reference point.

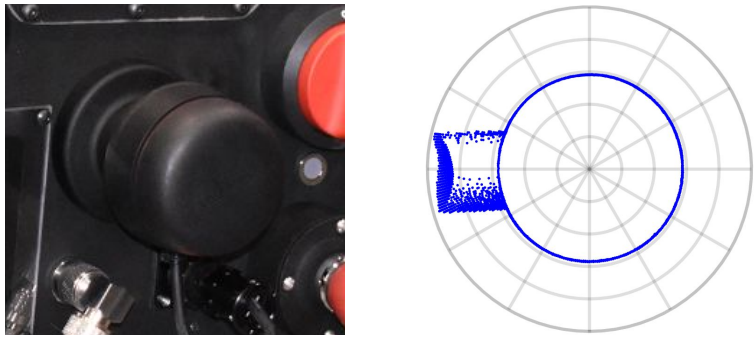

Fig. 6. (Left) RadPiper's geometric laser profiler. (Right) cross-sectional profile of a process pipe segment containing a vacuum port.

An inertial measurement unit (IMU, Xsens MTi-20) is mounted inside the robot body. The IMU is used to determine the orientation of the robot with respect to gravity. From a radiometric perspective, this can be used to flag segments for which assumptions on the position of the robot relative to the pipe are broken due to excessive pitch. RadPiper also uses its IMU to actively steer along the axis of the pipe, maintaining the centering assumed in the radiometric model. A final use is for safeguarding, for instance, if RadPiper drove into a reducer fitting, it would exceed its pitch threshold and automatically reverse.

RadPiper wirelessly transmits data from all three of these sensors (camera imagery, geometric cross-sections, and robot orientation) to its paired NDA technician tablet interface in real time while in communications range (tested out to 100 feet).

## STRUCTURES AND MECHANISMS

A core benefit and novelty of the RadPiper robotic crawler in PCAMS is its ability to assay both 30- and 42-inch process pipes, providing NDA operators with spares and fleet reliability throughout high-cadence D&D schedules. This capability is realized through the manual transformation of any robot's collimator assembly to center the gamma detector along the axis of a 30- or 42-inch pipe and tilt the tracks to be tangent to the respective pipe wall. Fig. 7 illustrates this transformation.

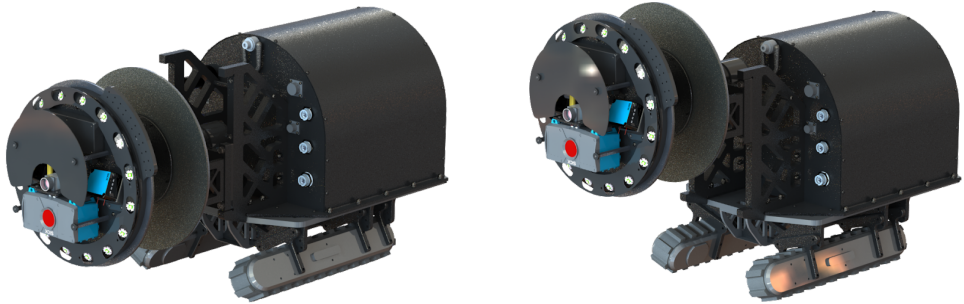

Fig. 7. RadPiper in (left) 30- and (right) 42-inch configurations; note the raised collimator height and decreased track angle.

In addition to centering, the collimated detector is cantilevered from the front of the robot structure, ensuring the detector's field of view does not impinge on the robot or its potentially contaminated tracks. The sensor suite on the front of the robot is also kept forward and centered in the pipe as it includes the fisheye camera and forward safeguarding mapper. Wires for these sensors are guided across the collimator by a plastic guide at a constant radius from the detector crystal and a varying clocking angle in order to minimize the attenuation effect of this impingement on the gamma spectroscopy. Sensors on the rear of the robot, particularly the laser rangefinder and geometric profiler, must be rigidly mounted to its frame due to the importance of their relative positions and orientations for localization and geometric modeling. These sensors are also modularly removable to facilitate change-out in the case of sensor failure.

All external sensors and the robot's wireless antenna are removable by technicians in Personal Protective Equipment (PPE) without opening the robot's main arched compartment, which can be opened but is sealed against contamination. RadPiper's removable USB jump drive is contained in a separate resealable screw-top compartment which is readily accessible by operators in full Powered Air Purifying Respirator PPE. Robot handling and maintenance operations related to these features are discussed operationally in [7].

Along with these sealing features, the entire RadPiper robotic platform is designed in support of automated NDA for defunct gaseous diffusion facilities. The anodized structure is fully conducive to wipedown, and the mobility platform detaches with six bolts and can be fully submerged for decontamination. Each track can also be removed and submerged, though thus far contamination has been minimal enough to require only cutting off and replacing the rubber treads of the tracks. The front sensor suite, which is necessarily exposed for robot safeguarding, is fully modular and can be replaced for sensor switch-out or separate decontamination.

## ELECTRONICS AND COMPUTING

RadPiper's electronics architecture integrates industrial-grade hardware sealed within a single enclosure with removable connections to all external sensors and components. It is designed to maintain reliability while operating autonomously in contaminated process piping and does not require internal access by NDA technicians in the field.

To facilitate deployment and inspection operations, RadPiper is powered by internal lithium iron magnesium phosphate batteries rather than via a tether as is common in other pipe inspection robots [9,10,11]. The batteries include integrated battery management and power the robot for at least a full NDA shift, a total of five hours of continuous operation. Charging time of approximately two hours allows rapid turnaround of robot service. Safety lockouts ensure the robot cannot be driven while connected to the charger.

Because the electronics module is fully sealed, internal temperature is regulated via an internal fan to mitigate hot spots and via heat sinking key components to the external walls of the enclosure. RadPiper's computer, a quad core Intel i7, is custom-machined to be thermally grounded on the front bulkhead, and the track motor drivers and voltage converters are similarly grounded to the bottom plate of the enclosure. Temperature sensors located on key components automatically trigger an operator alert and, if disregarded, trigger robot reversal upon reaching thermal thresholds. RadPiper can operate continuously in ambient temperatures up to 100°F (42°C). Fuses protect batteries, DC to DC converters and drivers. The LED light ring on the front of the robot is also mounted to a custom heatsink and the driver synchronized with camera frame capture to minimize power expenditure.

Reliability of RadPiper's key sensors (gamma detector, safeguarding mapper, laser rangefinder, inertial measurement unit, and camera) is enhanced through direct linking to the robot computer, as is the data logging USB drive. Left and right track motions are coordinated under CANopen (controller area network) protocol. Both track controllers also connect directly to rear panel manual jog and motion stop switches, ensuring robustness by avoiding reliance on the main computer for these critical controls.

A simplified schematic of RadPiper's electronics architecture is shown in Fig. 8.

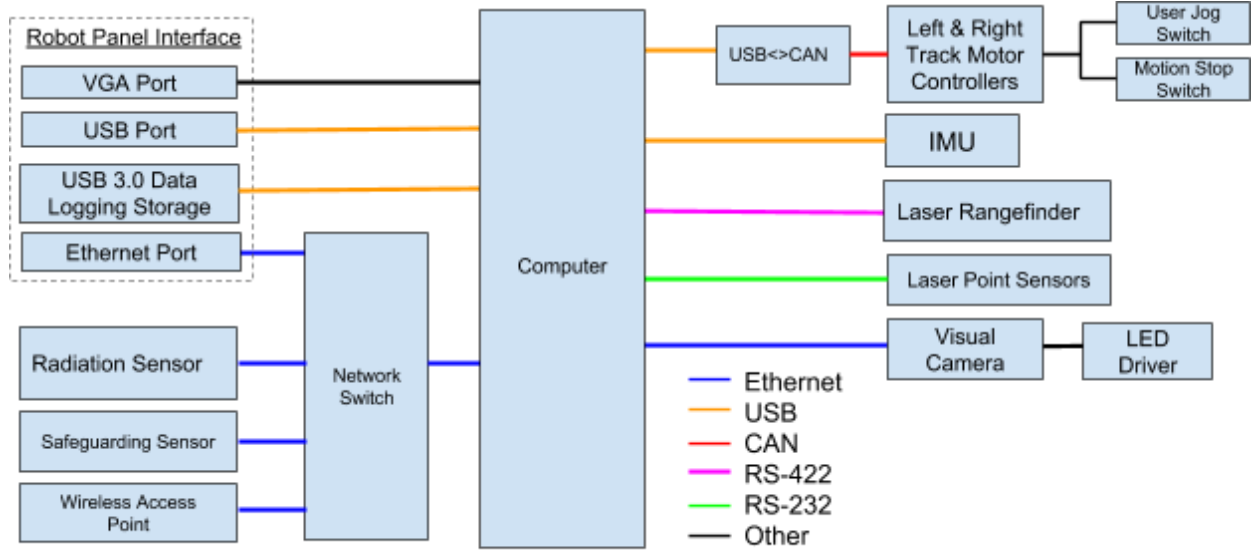

Fig. 8. RadPiper electronics architecture.

## AUTONOMY

### Robot Executive Controller

At the heart RadPiper's onboard autonomy is its robot executive, programmed to prioritize operational safety by ensuring the failsafe condition is safe return of RadPiper to its launch rig. This safety prioritization is implemented through a finite state machine that allows for unambiguous verification of each state of operation.

At the start and end of every run, RadPiper enters a quality control (QC) mode. In this state, the robot verifies everything from radiation data to low-level device operation. The executive ensures all physical characteristics of the robot meet requirements. This includes proper temperatures and voltage levels at different parts of the internal circuitry. It also ensures that the robot meets non-hardware requirements: all software nodes must be in good health and there must be enough space to store all required data.

Once RadPiper completes QC and begins motion, the executive verifies proper operation of all safety-critical components. It cross-checks multiple sensors to ensure not only that all errors are detected, but also that no false errors are raised. This method is primarily used for onboard localization of the robot. The rear-facing laser rangefinder is generally very accurate, but when the robot drives a long distance down the pipe, the laser can lose sight of its launch rig target, causing the reading to fluctuate wildly. Looking at this data alone makes it difficult to ensure that the robot is travelling at the proper speed in the proper direction. Cross-checking this data with the IMU and the encoders provides confidence in operation.

While traveling forward, there are three conditions for reversal: the pre-specified distance is reached, safeguarding indicates untraversable terrain, or the executive detects a failure in a hardware or software component. Upon reversal and return to its start location, the robot uses the various localization methods to return to the same exact position on its launch rig for operational safety and direct comparison to QC and contamination checking. This reinforces the core requirement that robot autonomy bring RadPiper back to its start condition at all costs. Disabling of a robot inside the enrichment cascade would result in costs and schedule delays that undermine confidence in robotic NDA.

**Safeguarding**

Robot safeguarding utilizes three independent sensors to ensure RadPiper does not attempt to traverse an impassable obstacle, be it a holdup deposit, pipe fitting, or cut pipe section, in either forward or reverse. (Reverse traversability is verified before continuation of the forward run.) The primary sensor is an infrared time-of-flight mapper for which the robot processes a point cloud in real time to detect different reversal criteria. This is the most versatile but also the most complex sensing and autonomy method, and thus is complemented by two single-point laser rangefinders and the robot's IMU.

RadPiper's safeguarding infrared mapper creates a depth image of the environment in front of the robot at a 50 Hz sampling rate. The resulting point cloud is pre-filtered to include only high-confidence ranges, which reduces computation and response time. The filtered point cloud is then modeled via random sample consensus (RANSAC) [12] to estimate parameters of a mathematical pipe cylinder model. This cylinder model is used to estimate the robot's six-degree of freedom pose with respect to the center of the pipe.

Reversal conditions are assessed by segmenting the point cloud into three-dimensional sections that include points conducive with the cylinder model and points that are not (Fig. 9). There are three geometric checks for geometric end-of-run conditions. First, RadPiper checks for the presence of an impassable obstacle by analyzing the non-cylindrical point cloud points for specific RadPiper traversal capabilities (e.g. height of obstacle under the tracks). Second, RadPiper sweeps a plane along the pipe axis and detects any using a Hough Transform [13]. Filled circles are treated as closed pipe (e.g. a closed gate valve) conditions, decreasing diameter below a threshold is treated as reducer, and a loss of open circles indicates an open cut pipe end. If RadPiper's algorithm detects either an in-pipe obstacle or an open or closed end of a pipe, it calculates the distance to this condition and passes it through a finite impulse response (FIR) low pass filter to denoise the signal. It then reports this end-of-run condition and the denoised distance to it to the robot executive, causing RadPiper to enter Approach Mode to cover this distance and inspect the last segments of pipe. This method has been demonstrated successfully on reducers, swept-T joints, and closed valves in 30- and 42-inch process piping.

The secondary safeguarding method uses much simpler single-point laser triangulation sensors mounted at compound 26-degree angles to point in front of the robot's tracks. Knowing the angular state of the robot (from its IMU) and the distance at which the point sensors should measure the 30- or 42-inch pipe wall at all times allows thresholding of these distance measurements (after denoising). Shorter measurements indicate a large obstacle under a track, or a steep pipe fitting. Larger measurements indicate a hole in the pipe. This method is very simple and less susceptible to unknown pipe geometries and sensor and robot failures. It has override authority if the primary three-dimensional safeguarding algorithm has not detected a reversal condition. The third and last safeguarding sensor is the RadPiper IMU itself, which has thresholds for minimum and maximum robot pitch (e.g. driving into a reducer or open valve) and can similarly override the primary safeguarding algorithm.

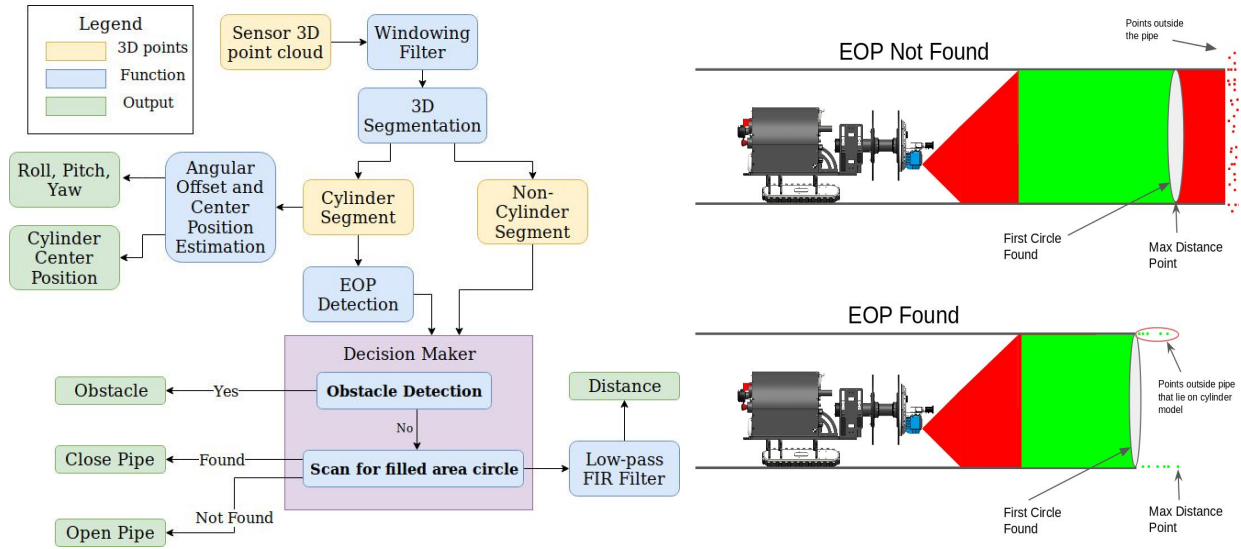

Fig. 9. (Left) flowchart of the safeguarding algorithm. (Right) end of pipe (EOP) detection illustration.

**LOCALIZATION**

Once RadPiper successfully inspects and exists a process pipe, post-processed reporting of its radiometric sensing relies on precise localization of these sensor readings. This localization method relies on three robot sensors: the geometry profiler for determining the start of the pipe, the laser rangefinder for absolute distance measurements, and the track encoders for interpolation between absolute rangefinder readings on long pipe runs.

RadPiper determines the entrance of each gaseous diffusion pipe by sensing the geometric transition between the larger space of the cell floor and the confined pipe diameter, as registered on its helical geometric profile map of each pipe. The geometric profiler begins spinning outside on the robot launch rig (illustrated in Fig. 10 left), allowing PCAMS to detect the transition into the pipe and set when the detector crystal crosses this point as the start of its first NDA segment. This is also the point from which operators measure sections of pipe for removal.

Localization beyond the entrance of the pipe is based primarily on RadPiper's single-point laser rangefinder. This sensor, shown in Fig. 10 right, continuously records distance from the robot back to the target on its launch rig while traversing a pipe. This provides the robot's absolute location once adjusted for the position of the pipe entrance as registered by the geometric profiler.

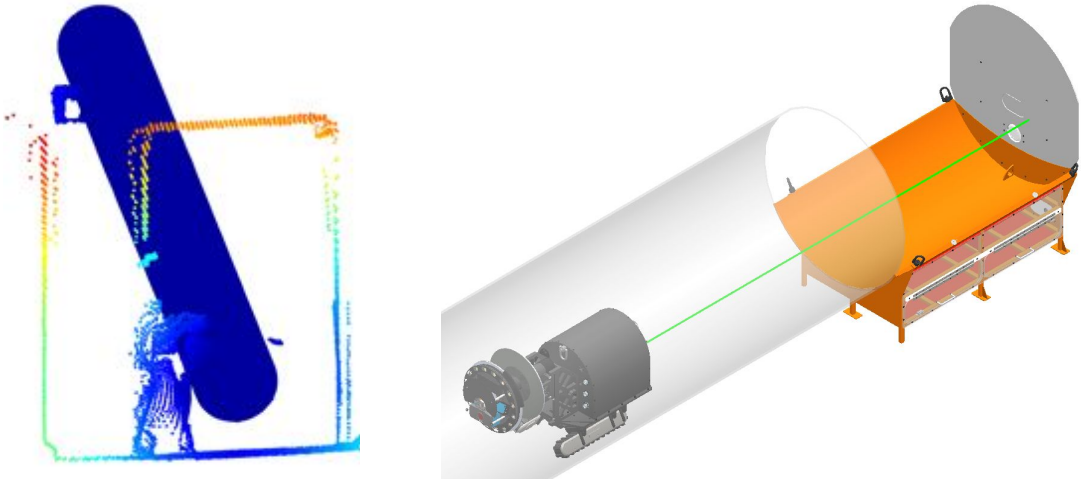

Fig. 10. (Left) the geometric point cloud created by RadPiper's laser profiler, with the exterior cell floor and robot operator visible in the foreground. (Right) Render of the RadPiper laser rangefinder measuring distance the back to the robot's launch rig with the laser illustrated in green.

For long pipe runs, in-pipe artifacts and steering geometry may preclude continuous registration of the laser rangefinder to the launch rig at the pipe entrance. When this continuous absolute measurement is not available, localization is supplemented by incremental odometry from RadPiper's track encoders. Experimental deployment has shown consistent rangefinder readings up to 40-50 feet with intermittent fixes out to over 100 feet. These intermittent readings are used to adjust for encoder drift.

PCAMS uses factor graph optimization to fuse localization sensor data into the interpolated position and uncertainty of RadPiper's gamma detection field of view throughout forward and reverse traversals of pipe. Gamma detector spectra, adjusted track encoder counts, laser rangefinder distance measurements, and laser profiler registration of the pipe entrance all have known position transforms with respect to each other, and all data points are time stamped by RadPiper when collected. The method of fusing these data is factor graph optimization, a state-of-art graphical method for solving simultaneous localization and mapping (SLAM) problems. PCAMS PPS runs factor graph optimization to probabilistically localize each of RadPiper's sensor measurements along a forward and reverse pipe length. This same factor graph optimization method simultaneously returns combined localization uncertainty at each point. This localization uncertainty is minimal and based on the input uncertainty characterization of each sensor as determined by testing under known ground truth conditions. Note that for both input sensor and factor graph uncertainty, the only notable radiometric effect is the gross shifting of measurement location. While localization values do have some minimal effect on the PCAMS spectroscopic moving average (described in [5]), this effect is orders of magnitude below that of counting uncertainty and is addressed by conservatism in localization tolerances. A model factor graph illustration is shown in Fig. 11. A series of nodes (blue circles) representing RadPiper's forward and reverse pose time series are joined by several types of edges (lines with smaller colored circles) that represent sensor measurements between the pose states.

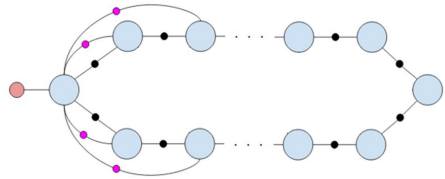

Fig. 11. Factor graph illustration of the PCAMS localization process. The large blue circles are a time series of robot poses. The lines with colored circles represent sensor measurements between these poses: odometry measurements (black), and laser rangefinder measurements (pink).

In PCAMS, factor graph edges (i.e., sensor measurements) are primarily absolute distance readings from the laser rangefinder. When the robot traverses far enough that these readings become intermittent, the optimization algorithm uses these intermittent fixes to adjust the incremental track encoder readings. Details of how this localization is integrated with per-foot reporting of U-235 loadings is included in [6].

## RESULTS OF PERFORMANCE TESTING

Before delivery and testing with known U-235 sources and contaminated piping at DOE Portsmouth, RadPiper robots underwent an extensive test regime to qualify its deployment tolerances, robot autonomy and safeguarding, and localization. Radiometric calibrations and tests are detailed in [5].

### Deployment Tolerance Tests

PCAMS guarantees operators the ability to deploy into pipes when the launch rig appears up to 2 cm high or low of the entrance, 2.5 cm left or right, 6 degrees yawed, and/or 5 cm backward. The cut edge of the pipe itself can be angled up to 5 cm. Deployment testing has verified these tolerances, including surmounting a 5 cm step and bridging an 18 cm gap in both 30- and 42-inch configuration. Fig. 12 shows images from some of these deployment tests, conducted at both high and low elevation.

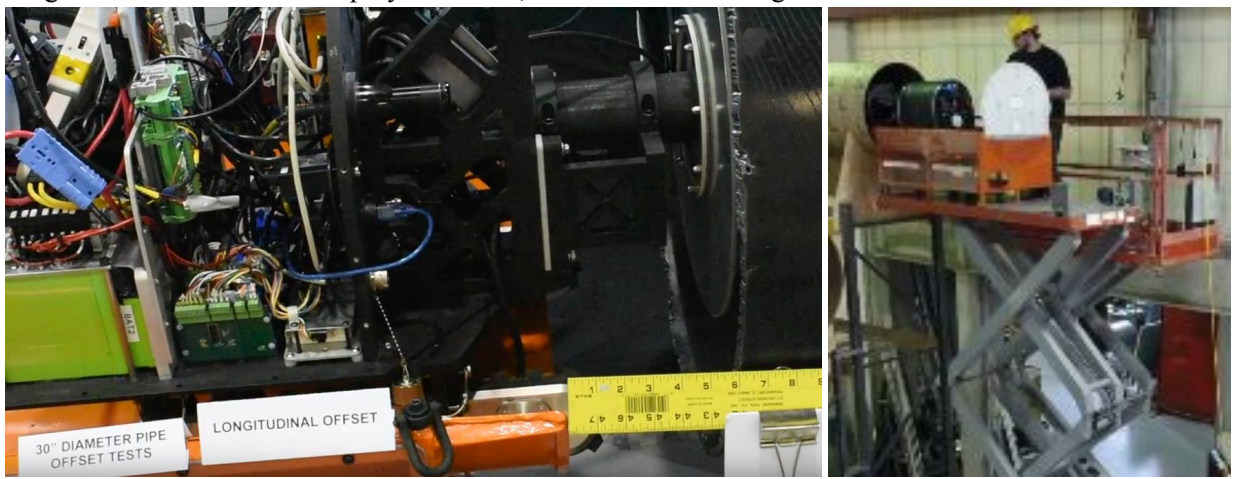

Fig. 12. Images from low and high deployment testing or RadPiper: (left) RadPiper crossing an 18 cm gap and (right) a PORTS technician deploying RadPiper into a high pipe at Carnegie Mellon.

### Safeguarding Tests

The RadPiper safeguarding algorithm was tested in mockups of open and closed pipe ends, obstacles, swept-T joints, and reducers. Following mockup qualification, Fig. 9 shows the algorithm approaching a swept-T in a Portsmouth process pipe. It registers the fitting at 8 seconds as a decrease in its steady-state 1.5m distance. The blue line indicates the denoised raw red data. RadPiper travels its Approach Mode distance and begins reversal at 9.5 seconds. The drift in this distance estimate is minimal: shown to be between 3 and 9 cm for all end-of-pipe cases over 20 runs.

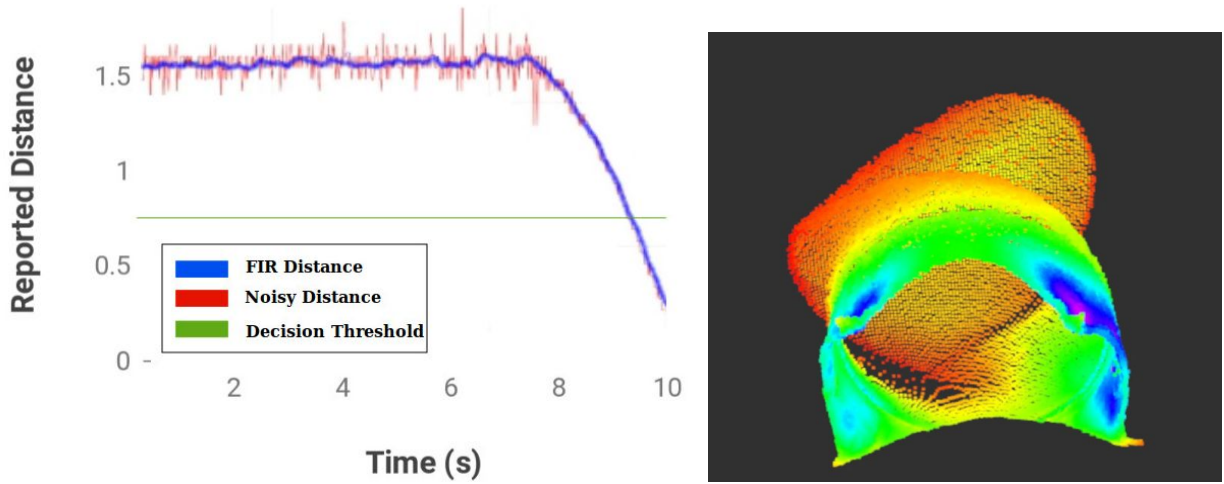

Fig. 13. Left: The reported distance of swept-T during hot test vs time. Right: swept-T point-cloud

**Localization Tests**

Long distance RadPiper localization was qualified by comparing repeated 100-foot (30-m) runs in a clean pipe at Carnegie Mellon to ground truth robot position tracked by a survey instrument. As RadPiper traverses farther into the pipe (here beyond 25 m), its laser rangefinder begins to hit sections of pipe wall rather than the launch rig target. This raw incorrect data is shown in Fig. 14 left. The filtered localization result from the PCAMS algorithm is shown on the right, closely matching the ground truth. As it is shown, localization results matches ground truth when rangefinder data is available, on the other hand, track odometry is the only source of localization when rangefinder data is discarded, hence, localization could drift a bit. However, as soon as rangefinder data becomes available again on the reverse run, localization results corrects itself. Fig. 15 shows error against the ground truth for one of our tests. Our results show max drift of 10 cm over a 33 m run.

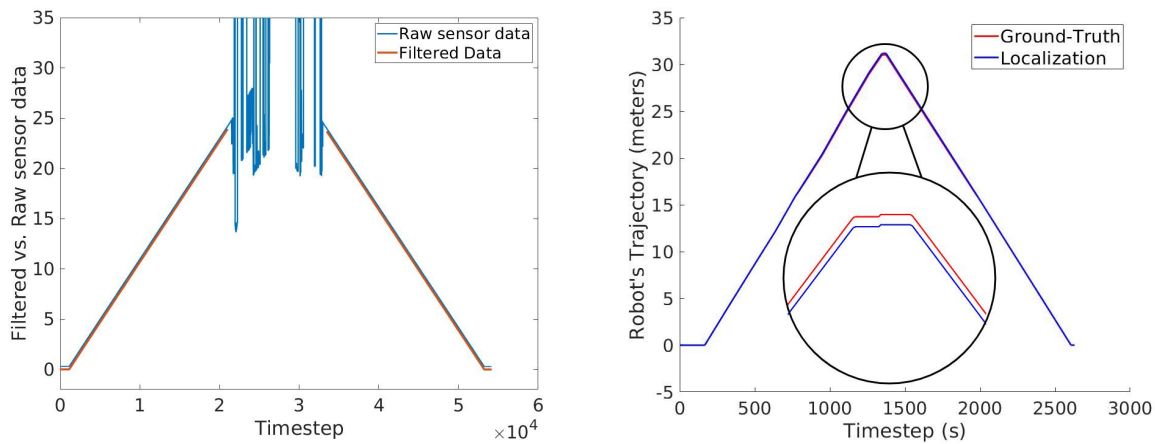

Fig. 14. Left: Filtered vs. Raw sensor data. Right: Localization vs. ground truth

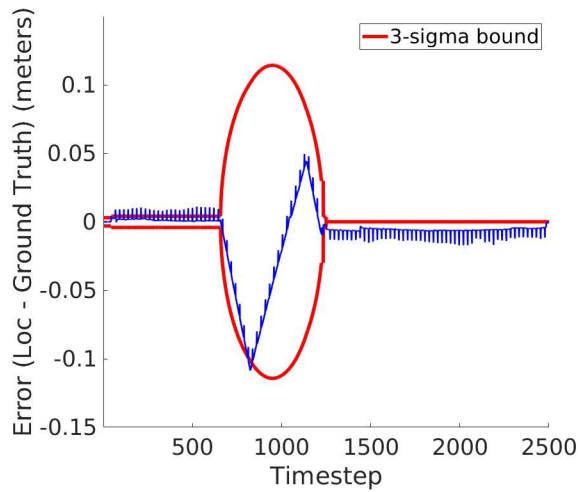

Fig. 15. Localization error results

**Integrated Tests**

The integrated RadPiper robot was tested for PORTS and DOE at an acceptance test at Carnegie Mellon's pipe inspection test facility in May 2018. The culmination of this test program was a mock Performance Demonstration Program (PDP) test using enclosed strips of Co-57 (U-235 is not available at the off-site facility). RadPiper exhibited the ability to traverse a 100-foot pipe forward, trigger its safeguarding on the open far end of the pipe, and traverse 100 feet backward to return onto its launch rig while maintaining constant onboard datalogging and autonomy as well as wireless communication with operators. Post-processing of the data confirmed that RadPiper correctly recorded radiometric, odometric, geometric, pose, and visual data as well as robot status and health. Fig. 16 presents samples of the robot's raw gamma, camera, and geometric profile data. Fig. 17 shows a sample of the localized radiometric readings versus the a priori unknown ground truth. Results of the full PCAMS system (including deployment system, analysis software, etc) and NDA method as tested with known sources and contaminated piping at PORTS are detailed in [5,7].

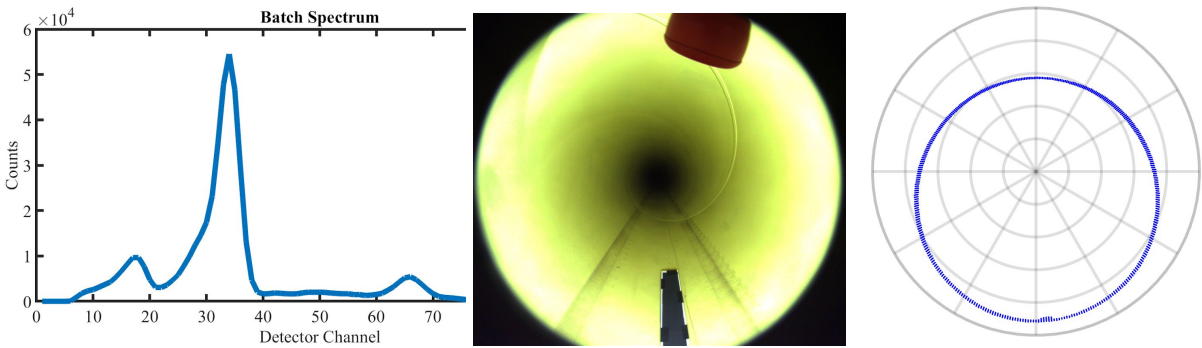

Fig. 16. Samples of RadPiper's raw (left) gamma, (center) camera, and (right) geometric data.

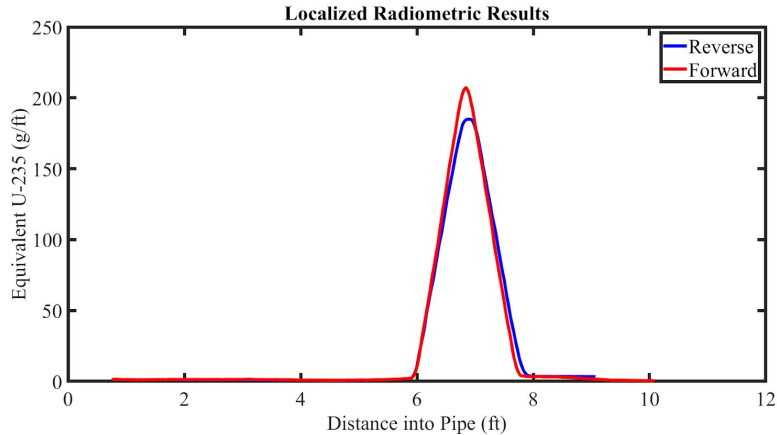

Fig. 17. Sample plot of localized radiometric results from forward (red) and reverse (blue) traversals.

**CONCLUSIONS**

Two production prototypes of the new PCAMS RadPiper robot for measuring U-235 deposits in pipes have been designed, constructed, and tested. Key advances over prior work include onboard quality control checking, ability to transform for two pipe sizes, advanced autonomy to handle a wide range of pipe fittings and conditions, and increased precision and robustness of localization.

Tests have exhibited that the robot can be launched into low and high elevation pipes at 30" and 42" diameters. The step, angle offset, and gap crossing limits have been characterized and exceed required targets. Robot autonomy and safeguarding operate as expected in repeated runs with a variety of pipe conditions. Localization tests show positioning accuracy to 10cm over a 33 m run. Results from integrated testing with Co-57 sources were also presented showing repeatability in counting efficiency and localization between forward and reverse traversals.

Future work includes automatic identification of pipe features such as expansion joints from within pipes and a remote operator override for detected obstacles. Robots for smaller diameter pipes, down to 3 inches (7.6 cm), are also in development. This presents significant miniaturization challenges for all components and requires development of a radiometric approach for a smaller scintillation detector.

The development of robotic NDA of holdup deposit has proceeded rapidly from a first proof-of-concept for 30" pipe in 2017 to robust production prototypes for 30" and 42" pipe in 2018. One PCAMS RadPiper robot has been delivered to the DOE Portsmouth site and is going through the commissioning process. In 2019, site contractors will conduct regular robotic measurements as part of site D&D, saving many hours of dangerous, elevated work and greatly accelerating NDA progress.

## ACKNOWLEDGEMENTS

Funding for this work was provided by the US Department of Energy under cooperative agreements DE-EM0004383 and DE-EM0004478.